\documentclass[letterpaper]{article}
\PassOptionsToPackage{numbers, compress}{natbib}

\usepackage[preprint]{neurips_2021}

\usepackage[utf8]{inputenc} 
\usepackage[T1]{fontenc}    
\usepackage{hyperref}       
\usepackage{url}            
\usepackage{booktabs}       
\usepackage{amsfonts}       
\usepackage{nicefrac}       
\usepackage{graphicx}
\usepackage{subfigure}
\usepackage{ulem} 
\usepackage{comment}
\usepackage{enumitem}
\usepackage{multirow}
\usepackage{amsmath}
\usepackage{amssymb}
\usepackage{microtype}      
\usepackage{xcolor}         
\usepackage{hyperref}

\usepackage[linesnumbered,ruled,vlined]{algorithm2e}

\title{MGAE: Masked Autoencoders for \\ Self-Supervised Learning on Graphs}

\author{%
  Qiaoyu Tan \\
  Texas A\&M University\\
  \texttt{qytan@tamu.edu} \\
   \And
   Ninghao Liu \\
   University of Georgia \\
   \texttt{ninghao.liu@uga.edu} \\
   \AND
   Xiao Huang \\
   The Hong Kong Polytechnic University \\
   \texttt{xiaohuang@comp.polyu.edu.hk} \\
   \And
   Rui Chen \\
   Samsung Research America \\
   \texttt{rui.chen1@samsung.com} \\
   \And
   Soo-Hyun Choi \\
   Samsung Research America \\
   \texttt{soohyunc@gmail.com} \\
   \And
   Xia Hu \\
   Rice University \\
   \texttt{xia.hu@rice.edu} \\
}

\begin{document}

\maketitle

\begin{abstract}
We introduce a novel masked graph autoencoder (MGAE) framework to perform effective learning on graph structure data. 
Taking insights from self-supervised learning, we randomly mask a large proportion of edges and try to reconstruct these missing edges during training. MGAE has two core designs. First, we find that masking a high ratio of the input graph structure, e.g., $70\%$, yields a nontrivial and meaningful self-supervisory task that benefits downstream applications.
Second, we employ a graph neural network (GNN) as an encoder to perform message propagation on the partially-masked graph. To reconstruct the large number of masked edges, a tailored cross-correlation decoder is proposed. It could capture the cross-correlation between the head and tail nodes of anchor edge in multi-granularity. Coupling these two designs enables MGAE to be trained efficiently and effectively. Extensive experiments on multiple open datasets (Planetoid and OGB benchmarks) demonstrate that MGAE generally performs better than state-of-the-art unsupervised learning competitors on link prediction and node classification. 

\end{abstract}

\section{Introduction}
\label{intro}
Graph structure data is ubiquitous in real-world systems~\cite{hamilton2017representation,chen2020graph}, such as social networks, academic graphs, and biological interaction networks. Given that labels are often not available, unsupervised graph representation learning has attracted considerable attention in both academia and industry. The goal is to learn node representations to preserve the input graph structure~\cite{perozzi2014deepwalk,kipf2016variational,pan2018adversarially,tan2019deep}. Based on the learned representations, we could perform various unsupervised tasks, such as link prediction and anomaly detection. Also, we could directly apply off-the-shell learning algorithms to the learned representations to perform supervised tasks, such as node classification.


To learn node representations in a unsupervised manner, there are two lines of research. First, graph autoenocders, such as GNN encoder based autoenocders, are proved to be effective in many graph domains~\cite{zhang2018network,zhang2020revisiting} on node classification and link prediction tasks. 
It aims to reconstruct the original network structure (i.e., observed edges) for model training. Efforts have been devoted to exploring effective encoder networks. For example, 
GAE~\cite{kipf2016variational} adopts the classical graph convolutional network (GCN)~\cite{kipf2016semi} model as encoder, while GraphSage~\cite{hamilton2017inductive} introduces an inductive variant of GCN for graph encoding. 
Second, graph self-supervised learning (GSSL) focuses on designing advanced pretext tasks for self-supervised training~\cite{jin2020self,xu2021infogcl}. For example, DGI~\cite{velivckovic2018deep} and GIC~\cite{mavromatis2021graph} target to train GNN models by maximizing the mutual information~\cite{bachman2019learning} between the node-level representation and the graph-level representation where the anchor node located in. 
GSSL methods could learn robust and powerful GNN models for graph encoding~\cite{liu2021self}. 



While edge reconstruction and edge dropping are common techniques in graph autoenocders and GSSL, masked autoencoding has never been explored for graphs. Its key idea is to remove a proportion of the input data and use the removed content to guide the traning. Masked autoencoding has been proven to be effective and efficient in modeling texts~\cite{devlin2018bert} and images~\cite{he2021masked}. Graph autoenocders take the complete graph as input and target at reconstructing the entire edges. In GSSL, one of the graph augmentation methods is to drop some edges~\cite{jin2020self}. Edge dropping may not work well in many scenarios as shown in experiments. Its goal is to learn robust representations, instead of predicting the removed edges.  Besides, in practice, its valid masking ratio value is less than $30\%$. Therefore, we ask: \textit{How to design appropriate masked autoencoding for graphs? What percentage of edges are really necessary to reconstruct the input graph and learn effective node representations}?



\begin{figure*}[t]
\begin{center}
\includegraphics[width=14 cm]{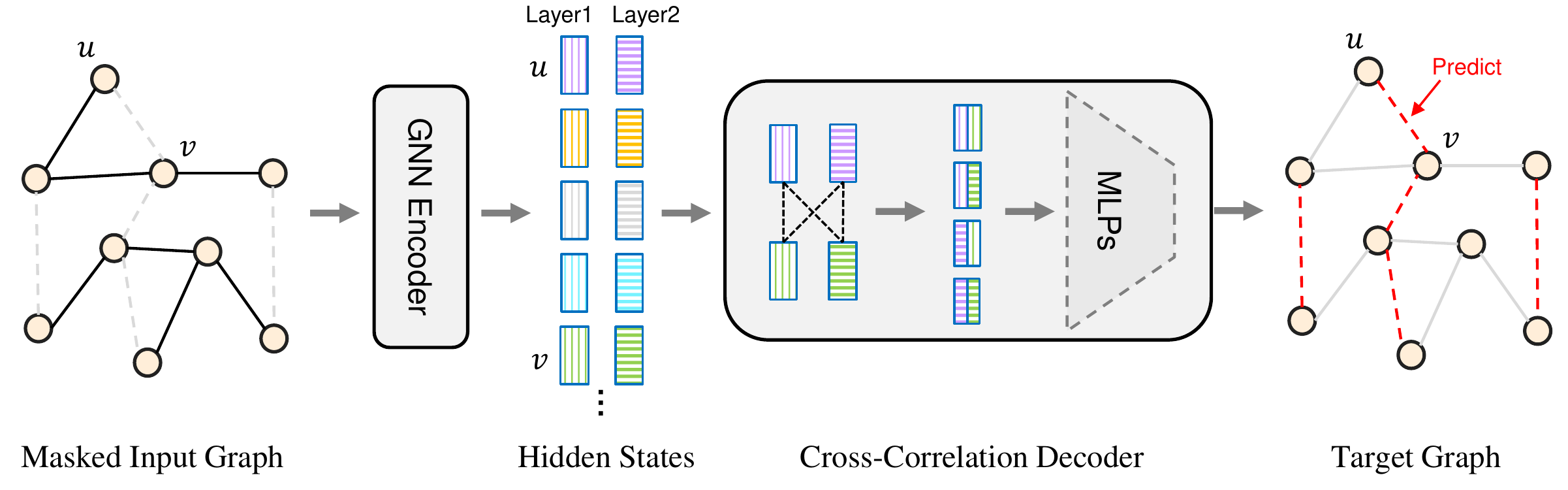}
\caption{\textbf{The proposed MGAE architecture}. Given a graph, a large random subset of edges are masked. The GNN encoder is applied to the remaining edges to produce hidden representations. The representations are then fed into a tailored decoder to reconstruct the masked edges during training, which captures the cross-correlation between end nodes with multi-granularity features. }
\label{fig:mgae}
\end{center}
\end{figure*}

In this paper, we give a positive answer to these open questions. We present a simple yet effective graph autoencoder framework named masked graph autoencoder (MGAE) for unsupervised graph representation learning. MGAE targets to randomly mask a large proportion of the input graph structure and then recover the masked edges. Different from traditional graph autoencoders, I) our GNN encoder operates only on partial network structure (without masked edges) for convolution, and II) our decoder is designed to capture the cross-correlation between the head and tail nodes of an anchor edge to effectively reconstruct the link from their latent representations (See Figure~\ref{fig:mgae}). Under this design, our MGAE model can achieve a win-win scenario with a significant high masking ratio (e.g., $70\%$): it optimizes model performance while allowing the GNN encoder to process only a small portion ($30\%$) of original graph structure. We summarize our main contributions:
\begin{itemize}
    \item We introduce a novel graph autoencoder alternative, termed as masked graph autoencoder (MGAE), for graph-structured data. It is inspired by self-supervised learning and is not only robust and effective but also well suited for link prediction and node classification. 
    \item We propose a tailored cross-correlation decoder to effectively utilize the noisy hidden representations incurred by the masked graph structure for edge reconstruction. This meticulous design allows us to perform a very high masking ratio (e.g., 70\%) for edge masking, which also improves the effectiveness and efficiency.  
    \item Extensive experiments demonstrate that MGAE performs better or sometimes on par with state-of-the-art competitors, across Planetoid and OGB bechmarks in terms of link prediction and node classification tasks. 
\end{itemize}

\section{Problem Statement}
\textbf{Notations:} We use boldface lowercase letters (e.g., $\mathbf{x}$) to denote vectors, and boldface uppercase letters (e.g., $\mathbf{X}$) to denote matrices. The $v^{\text{th}}$ row of $\mathbf{X}$ is represented as $\mathbf{x}_v$. We assume an undirected graph $\mathcal{G}=(\mathcal{V}, \mathcal{E})$ with $n$ nodes is given, where $\mathcal{V}$ and $\mathcal{E}$ denote the sets of nodes and edges, respectively. Each node $v\in\mathcal{V}$ has a $F$-dimensional attribute vector $\mathbf{x}_v\in\mathbb{R}^F$ describing its properties. When node attributes are not available, $\mathbf{x}_v$ can be initialized as a one-hot index vector or learnable parameters. To study graph representation learning under the autoencoder framework, we follow the literature~\cite{kipf2016variational,pan2018adversarially,salha2019degeneracy} introduced as below. 

\textbf{Graph Autoencoder (GAE).} Given a graph $\mathcal{G}=(\mathcal{V}, \mathcal{E})$, the goal is to learn an encoder network $f: \mathcal{V}\times \mathcal{E}\xrightarrow[]{} \mathbf{H}$ that maps each node $v\in\mathcal{V}$ into a $d$-dimensional embedding vector $\mathbf{h}_v\in\mathbb{R}^d$, and a decoder network $q: \mathbf{H}\xrightarrow[]{}\mathcal{E}$ that reconstructs the network structure (e.g., edges in $\mathcal{E}$) from the latent space $\mathbf{H}$. It is an unsupervised framework, where the key is to preserve the topological structure and node attributes in the latent space $\mathbf{H}$ through accurately recovering the observed edges in $\mathcal{E}$. The performance of graph autoencoder is then evaluated by link prediction and node classification tasks. 

\textbf{Encoder.} The encoder network is instantiated as GNNs models~\cite{kipf2016semi,hamilton2017inductive}, which are well-established representation learners on graphs. Following the message passing strategy~\cite{gilmer2017neural}, the core idea of GNNs is to update the representation of each node by aggregating itself and its neighbors' representations. Formally, at the $k$-th layer, we have, 
\begin{equation}
\mathbf{h}^{(k)}_v=\text{COM}(\mathbf{h}^{(k-1)}_v, \text{AGG}(\{\mathbf{h}_{u}^{(k-1)}: u\in\mathcal{N}_v\})),
\label{eq:mps}
\end{equation}
where $\mathbf{h}_v^{(k)}$ denotes the embedding of node $v$ at the $k$-th layer, and $\mathcal{N}_v=\{u| u | e_{v,u}\in\mathcal{E}\}$ is set of direct neighbors for node $v$. We often initialize $\mathbf{h}_v^{(0)}=\mathbf{x}_v$ in practice. The function \text{AGG} denotes the neighborhood aggregator. It selectively aggregates features from neighbors via either learnable attention weights~\cite{velivckovic2017graph,vaswani2017attention} or fixed combination weights determined by the graph topology~\cite{kipf2016semi,wu2019simplifying}.
To update node $v$, another function \text{COM} is used to combine the aggregated neighbor information and its own node embedding from previous layer. For a GNN model with $K$ layers, there are $K$ node representations $\{\mathbf{h}_v^{(1)}, \mathbf{h}_v^{(2)}, \cdots, \mathbf{h}_v^{(K)}\}$ being generated, where $\mathbf{h}_v^{(k)}$ captures the neighborhood structure within $k$ hops. 

\textbf{Decoder.} To reconstruct the observed edges in $\mathcal{G}$ from the hidden space $\mathbf{H}$, the decoder network is defined as an edge-wise similarity function. We want to predict the probability $y_{v,u}$ that edge $e_{v,u}$ exists. For example, some efforts~\cite{kipf2016variational} estimate the likelihood via inner product, i.e., $y_{v,u}=\mathbf{h}_v^{(K)}\cdot\mathbf{h}_u^{(K)^\top}$, while some other work~\cite{pan2018adversarially,hu2020open} parameterize the similarity function with multi-layer perceptron, i.e., $y_{v,u}=\text{MLP}(\mathbf{h}_v^{(K)}, \mathbf{h}_u^{(K)})$.

\section{Masked Graph Autoencoder}
We elaborate the core idea of the proposed Masked Graph Autoencoder (MGAE) framework in Figure~\ref{fig:mgae}. It is a simple approach that reconstructs the original network structure given only partially observed edges. Like the traditional graph autoencoders, MGAE has a GNN encoder that maps each node into a latent embedding, and a decoder that recovers the original links in the input graph from the latent space. The differences are as follows: (1) we allow the GNN encoder to operate only on partial network structure (i.e., a portion of edges are masked); (2) a novel cross-correlation decoder is designed to reconstruct the masked edges from the latent representation by capturing the cross correlation between the head and tail nodes of an edge in different granularity. There are four components in our approach: \textbf{network masking}, \textbf{GNN encoder}, \textbf{cross-correlation decoder}, and the \textbf{reconstruction target}. We will introduce the details of them as below. 

\textbf{Network masking.} We perturb the original graph $\mathcal{G}$ by randomly masking a subset of edges. Formally, we use $\mathcal{E}_{mask}$ and $\mathcal{E}_{reserve}$ to denote the edge sets that are masked out and remained, respectively, where $\mathcal{E}_{mask} \cup \mathcal{E}_{reserve} = \mathcal{E}$. To make the process more efficient, we adopt random sampling with a high masking ratio to obtain the masked edge set $\mathcal{E}_{mask}$. More sophisticated sampling strategies will be explored as future work. Specifically, we further consider two types of random sampling schemes to generate the masked input graphs. 
\begin{itemize}
    \item[I.] \textit{Undirected masking}. We treat the link between node $v$ and $v$ as undirected. That is, $e_{v,u}$ and $e_{u,v}$ are equivalent, and only one copy is included in $\mathcal{E}$. Therefore, performing random sampling over $\mathcal{E}$, the masked edge set $\mathcal{E}_{mask}$ is also undirected. 
    \item[II.] \textit{Directed masking}. The key assumption behind is that the links in the graph is directed. Hence, $e_{v,u}$ and $e_{u,v}$ are different, and both of them are included in $\mathcal{E}$. Deleting $e_{v,u}$ does not mean $e_{u,v}$ is also deleted. Therefore, after performing random sampling over $\mathcal{E}$, the resultant masked edge set $\mathcal{E}_{mask}$ is directed.
\end{itemize}
The main difference between the two masking schemes is that the undirected masking will make the reconstruction more challenging than the directed one. According to self-supervised principles, a more difficult task (e.g., masking a high ratio of edges) is expected to achieve better performance in theory. However, in network masking scenarios, we found that the best sampling strategy should be related to the network structure. For example, if the network is dense, it is better to adopt the tougher strategy (i.e., undirected masking). In contrast, if the original graph is sparse, the directed masking could be the better choice, since undirected masking may significantly destruct the neighborhood structure of nodes, leading to substantial expressive power degradation of GNN encoders. For example, in experiments, we found directed masking works better for OGB datasets, while undirected masking works better for Planetoid datasets.

\textbf{GNN encoder.} We follow standard GAE approaches to adopt the well-established GNN models shown in Eq.\eqref{eq:mps} as our encoder. Specifically, we consider GCN~\cite{kipf2016semi} and GraphSage~\cite{hamilton2017inductive} architectures as backbones. However, our encoder only operates on a small subset (e.g., 30\%) of the full edge set $\mathcal{E}$ for message propagation. The edges in $\mathcal{E}_{mask}$ are removed during training. This allows us to train the GNN encoder with reduced computation and memory costs. The masked edges are recovered by a tailored decoder, which will be introduced later.
It is worth noting that, with the same masking ratio, the total number of edges being used for training is the same for both sampling strategies discussed above. This is because the message propagation in GNNs is bidirectional by default. 
%

\textbf{Cross-correlation decoder.} The proposed MGAE decoder receives $K$ hidden representation matrices $\{\mathbf{H}^{(1)}, \mathbf{H}^{(2)}, \cdots, \mathbf{H}^{(K)}\}$ generated by the GNN encoder. See Figure~\ref{fig:mgae}. Each node $v$ has $K$ embedding vectors denoted by $\{\mathbf{h}_v^{(k)\in\mathbb{R}^d}\}_{k=1}^K$, where $\mathbf{h}_v^{(k)}$ is obtained by aggregating messages from neighbors of $v$ within $k$ hops defined by $\mathcal{E}_{reserve}$. Given that we mask a high ratio of edges in the original graph, the remained network structure (the adjacency matrix) spanned by the remained edge set $\mathcal{E}_{reserve}$ is inevitably incomplete. Directly using the final-layer representation $\mathbf{H}^{(K)}$ to reconstruct the masked edges is rather difficult. To tackle the challenge, we resort to modeling the correlation between two nodes in different granularities. Specifically, given an edge $e_{v,u}$ and their $K$ hidden representations $\{\mathbf{h}_v^{k}, \mathbf{h}_v^{(k)}\}_{k=1}^K$, we aim to capture their cross-correlations as below. 
\begin{equation}
\mathbf{h}_{e_{v,u}}=||_{k,j=1}^K\mathbf{h}_v^{(k)}\odot\mathbf{h}_u^{(j)},
\label{eq:cross}
\end{equation}
where $||$ denotes concatenation and $\odot$ denotes the element-wise multiplication. Here $\mathbf{h}_{e_{v,u}}\in\mathbb{R}^{dK^2}$ is the final representation for edge $e_{v,u}$. $\mathbf{h}_v^{(k)}\odot\mathbf{h}_u^{(j)}$ denotes the cross representation between node $v$ and node $u$, considering their $k$-th order neighborhood and $j$-th order neighborhood, respectively. The key insight behind the cross representation is to highlight the common patterns of the two nodes in different granularity features. This operation is crucial to MGAE because the $K$ latent embedding vectors themselves are incomplete with noise, so we have to identify their shared patterns as effective features for edge reconstruction. Additionally, we want to remark that $K$ is usually a small number (e.g., $K=2$) in graph autoencoders. Thus, the extra computation and memory costs are limited.

\textbf{Reconstruction target.} Different from traditional graph autoencoders, our MGAE recovers the original graph structure by learning to predict the masked edges $\mathcal{E}_{mask}$ rather than the observed ones in $\mathcal{E}_{reserve}$. This setting is inspired by recent self-supervised learning in computer vision~\cite{he2021masked}. By enforcing the MGAE decoder to only reconstruct the masked edges, our GNN encoder is robust to network noises and can encode graph information more effectively. We adopt the standard graph-based loss function to train our model. 
\begin{equation}
\begin{aligned}
\mathcal{L}=&-\sum\nolimits_{(v,u)\in\mathcal{E}_{mask}} \log\frac{\exp(\mathbf{y}_{vu})}{\sum_{z\in\mathcal{V}}\exp(\mathbf{y}_{vz})},
\end{aligned}
\label{eq_loss}
\end{equation}
where $\mathbf{y}_{v,u}=\text{MLP}(\mathbf{h}_{e_{v,u}})$ is the reconstructed score for edge $e_{v,u}$, and $\text{MLP}$ is a multilayer perceptron with ReLU activation function. In experiments, we adopt negative sampling~\cite{yang2020understanding,hu2021ogb} to accelerate the optimization, since the sum operation in denominator od Eq.\eqref{eq_loss} is computational prohibitive. Our MGAE model is optimized via stochastic gradient descent outlined in Algorithm~\ref{alg1}.

After training the MGAE model, similar to other graph autoencoder architectures, we can use the output of decoder to perform link prediction tasks on unseen edges. For node classification, we use the GNN encoder to generate representations for nodes in the graph. Then, we concatenate $K$ representations of each node as its final representation fed into the classifier. 

\begin{algorithm}[tb]
\DontPrintSemicolon
   \caption{Masked Graph Autoencoder (MGAE)}
\label{alg1}
   \KwIn{Graph $\mathcal{G}=(\mathcal{V},\mathcal{E})$, GNN encoder depth $K$, masking ratio $\omega$, embedding dimension $d$;}
 
\While{not converged}{
    Randomly masking $\mathcal{G}$ with ratio $\omega$ into two edge sets: $\mathcal{E}_{mask}$ and $\mathcal{E}_{reserve}$;\\
    Perform GNN encoding on the reserved edge set $\mathcal{E}_{reserve}$ according to Eq.~\eqref{eq:mps};\\
    Obtain the cross representations of edges in $\mathcal{E}_{mask}$ according to Eq.~\eqref{eq:cross};\\
    Model update by minimizing the reconstruction loss over $\mathcal{E}_{mask}$ according to Eq.~\eqref{eq_loss};\\
   }
{\bf Return} The trained MGAE model.
\end{algorithm}

\begin{table*}[htbp]
\centering
  \caption{Dataset Statistics.}
  \begin{tabular}{c| c |c |c | c | c}
   \toprule
     Data&\# Nodes &\# Edges &\# Features &Split ratio &\# Classes\\
     \hline
     Cora &$2,708$ &$5,429$ &$1,433$ &$85/5/15$ &$7$\\
     CiteSeer &$3,312$ &$4,660$ &$3,703$ &$85/5/15$ &$6$\\
     PubMed &$19,717$ &$44,338$ &$500$ &$85/5/15$ &$3$\\
     ogbl-ddi &$4,267$ &$1,334,889$ &- &$80/10/10$ &$-$\\
     ogbl-collab &$235,868$ &$1,285,465$ &$128$ &$92/4/4$ &$-$\\
     ogbl-ppa &$576,289$ &$30,326,273$ &- &$70/20/10$ &$-$\\
     ogbn-arxiv &$169,343$ &$30,326,273$ &128 &$-$ &$40$\\
     ogbn-proteins &$132,534$ &$39, 561, 252$ &$8$ &$-$ &$112$\\
 \bottomrule
\end{tabular}
\label{dataset_stat}
\end{table*}

\section{Experiments}
We evaluate the performance of MGAE in a variety of open graph datasets. Specifically, we try to answer two questions. \textbf{Q1:} How effective is MGAE against the state-of-the-art models on link prediction and node classification? \textbf{Q2:} How does our model perform under different masking ratios?

\textbf{Datasets and experimental settings}. We evaluate the  prediction performance of MGAE on six benchmark graph datasets including three Planetoid datasets~\cite{sen2008collective} (Cora, Citeseer and Pubmed), and three OGB datasets~\cite{hu2020open} (ogbl-ddi, ogbl-collab, and ogbl-ppa). For node classification, we evaluate on five datasets including Cora, Citeseer and Pubmed and two OGB node classification datasets: ogbn-arxiv and ogbn-proteins. The data statistics are summarized in Table~\ref{dataset_stat}.

Our model is built upon Pytorch~\cite{paszke2019pytorch} and PyG (PyTorch Geometric) library~\cite{fey2019fast}. We train MGAE for 200 epochs with Adam~\cite{kingma2014adam} optimizer and early stopping with a patience of 50 epochs. There are three hyper-parameters in our model, i.e., masking ratio $\omega$, embedding dimension $d$, and encoder layer $K$. We set $K=2$ and $\omega=0.7$ by default if not specified. For embedding dimension, we fix $d=128$ and $d=256$ for Planetoid and OGB datasets, respectively. Besides, we apply \textit{undirected masking} for Planetoid datasets and \textit{directed masking} for OGB datasets as default. 

\begin{table*}[t]
\caption{Link prediction results on Planetoid data with a masking ratio $0.7$. The best results are highlighted.} 
 \begin{small}
\setlength{\tabcolsep}{2pt}
{
\begin{tabular}{l cc cc cc cc}
\toprule
  &\multicolumn{2}{c}{Cora}
 &\multicolumn{2}{c}{CiteSeer}
 &\multicolumn{2}{c}{PubMed}\\
\cmidrule(r){2-3} \cmidrule(r){4-5} \cmidrule(r){6-7}
 &AUC &AP &AUC &AP
&AUC &AP\\
\cmidrule(r){1-3} \cmidrule(r){4-5} \cmidrule(r){6-7}
DGI &$90.02 \pm 0.80 $ &$90.61 \pm 1.00 $ &${95.53\pm 0.40}$ &${95.72\pm 0.10} $ &$91.24 \pm 0.60 $&$92.23 \pm 0.50$\\
GIC &${93.54 \pm 0.60} $ &${93.33 \pm 0.70}$ &$\bf{97.04\pm 0.50}$ &$\bf{96.80\pm 0.50} $ &$93.71 \pm 0.30$&$93.54 \pm 0.30 $\\
ARGE &${92.40 \pm 0.00}$ &${93.23 \pm 0.00}$ &$91.94\pm 0.00 $ &$93.03\pm 0.00 $ &${96.81} \pm 0.00$&${97.11 \pm 0.00}$\\
GAE &$91.09 \pm 0.01 $ &${92.83 \pm 0.03} $ &$90.52\pm 0.04 $ &$91.68\pm 0.05 $ &$96.40 \pm 0.01 $&${96.50 \pm 0.02}$\\
SAGE &$86.33 \pm 1.06$ &$88.24 \pm 0.87$ &$85.65\pm 2.56$ &$87.90\pm 2.54$ &$89.22\pm0.87$&$89.44\pm0.82$\\
SelfTask-GNN &$90.65\pm0.50$ &$91.55\pm0.93$ &$89.98\pm0.45$ &$91.52\pm0.52$ &${97.48\pm0.12}$ &${97.11\pm0.16}$\\
\midrule
MGAE-GCN &${93.52\pm0.23} $ &$\bf{94.46\pm0.24} $ &$93.29\pm0.49 $ &$93.81\pm0.40$ &$\bf{98.45\pm0.03}$ &$\bf{98.22\pm0.05}$\\
MGAE-SAGE &$\bf{95.05 \pm 0.76}$ &$\bf{94.50 \pm 0.86}$ &${94.85\pm0.49}$ &${94.68\pm0.34}$ &${97.38\pm0.17}$ &${97.11\pm0.19}$ \\
\bottomrule
\end{tabular}}
 \end{small}
\label{table_lp_planetoid}
\end{table*}

\begin{table*}[t]
\caption{Link prediction performance on OGB datasets with a masking ratio $0.7$, excepting ogbl-ppa with masking ratio $0.4$. Best results are highlighted. "OOM" means out of memory on a GeForce RTX 3090 GPU device (24GB). }
 \begin{small}
\setlength{\tabcolsep}{2pt}
{
\begin{tabular}{l cc cc cc cc}
\toprule
  &\multicolumn{2}{c}{ogbl-ddi}
 &\multicolumn{2}{c}{ogbl-collab}
 &\multicolumn{2}{c}{ogbl-ppa}\\
\cmidrule(r){2-3} \cmidrule(r){4-5} \cmidrule(r){6-7}
 &Hits@20 &Hits@30 &Hits@50 &Hits@100
&Hits@10 &Hits@50\\

\cmidrule(r){1-3} \cmidrule(r){4-5} \cmidrule(r){6-7}
DGI &$13.87\pm4.81$ &$15.31\pm5.52$ &OOM &OOM &OOM &OOM\\
GIC &$10.56\pm6.77$ &$13.29\pm7.44$ &OOM &OOM &OOM &OOM\\
GCN &$37.07 \pm 5.07$ &$51.56 \pm 4.19 $ &$44.75\pm 1.07 $ &$52.30\pm 1.01 $ &$2.52 \pm 0.47$ &$\bf{10.82 \pm 1.04 }$\\
GraphSage &$53.90 \pm 4.74$ &$65.80 \pm 6.94$ &$\bf{54.63\pm 1.12}$ &$\bf{60.23\pm 1.20}$ &$1.87\pm0.67$ &$8.92\pm2.28$\\
ARGE &$20.43\pm4.66$ &$23.86\pm6.53$ &$28.39\pm2.51$ &$37.66\pm1.98$ &$0.41\pm0.26$ &$3.83\pm0.84$\\
SelfTask-GNN &$42.26\pm4.85$ &$50.67\pm6.02$ &$33.03\pm5.11$ &$42.22\pm7.14$ &$0.65\pm0.30$ &$4.26\pm1.27$\\
\midrule
MGAE-GCN &$\bf{65.91 \pm 3.50}$ &$\bf{75.02 \pm 2.26} $ &$\bf{54.74\pm 1.06} $ &$\bf{61.01\pm 1.18} $ &$ \bf{3.98 \pm 1.33} $&$ 9.97\pm 1.55$\\
MGAE-SAGE &$\bf{66.00 \pm 9.49}$ &$\bf{75.18 \pm 6.57} $ &$49.27\pm 0.96 $ &${55.44\pm 0.82} $ &$ {1.37\pm 0.38} $&$ {4.79\pm 0.16}$\\
\bottomrule
\end{tabular}
}
\end{small}
\label{table_lp_ogb}
\end{table*}
\subsection{Link Prediction}
We start by answering \textbf{Q1} based on link prediction tasks.

\textbf{Baselines}. We consider the following baselines methods. Three representative GAE models (GAE~\cite{kipf2016semi}, GraphSAGE~\cite{hamilton2017inductive}, and ARGE~\cite{pan2018adversarially}), and three self-supervised methods (DGI~\cite{velivckovic2018deep}, GIC~\cite{mavromatis2021graph}, and SelfTask-GNN~\cite{jin2020self}). For fair comparison, we adopt EdgeMask as the self-supervised signals for SelfTask-GNN in experiments. 
We aim to provide a rigorous
and fair comparison between different models on each dataset by using the same dataset splits and training procedure. To be specific, for Planetoid datasets (Cora, CiteSeer, and PubMed), following~\cite{kipf2016variational}, we randomly split all edges into three sets, i.e., the training set (85\%), the validation set (5\%), and the test set (10\%), and evaluate the performance based on AUC and Average Precision (AP) scores. For OGB datasets (ogbl-ddi, ogbl-collab, and ogbl-ppa), we follow~\cite{hu2020open} to split the datasets into three sets according to the split ratios summarized in Table~\ref{dataset_stat}, and evaluate their performance using Hit rate (Hits@N), where $N$ is the number of nodes recalled. For our model, we consider two variants: MGAE-GCN and MGAE-SAGE meaning that we use the GCN and GraphSage architecture to implement our GNN encoder.

Table~\ref{table_lp_planetoid} and Table~\ref{table_lp_ogb} report the averaged results of 10 runs over Planetoid and OGB datasets, respectively. Jointly comparing the results on two tables, we have the following major observations. 
\begin{itemize}
    \item Our model MGAE performs better than graph autoencoder based baselines (AGRE, GAE, and GraphSage) on six datasets in almost all cases. Specifically, MGAE achieves comparable results with the best graph autoencoder baselines on ogbl-collab and ogbl-ppa datasets, while significantly outperforms them on other four datasets with great margin. In six datasets, MGAE obtains new state-of-the-art performance on Cora, PubMed and ogb-ddi datasets. 
    \item Compared with self-supervised learning baselines (DGI, GIC, and SelfTask-GNN), our model MGAE achieves substantial performance gains on five out of six datasets. In particular, MGAE only loses to GIC on CiteSeer dataset while outperforms on other five datasets.  Besides, we also observe that the performance gap between our model and three self-supervised baselines increases on OGB datasets. This result shows that graph autoencoder architecture is more suitable for link prediction task on large-scale datasets. 
    \item Another important observation is that \textit{none} of our two variants MGAE-GCN and MGAE-SAGE can consistently outperforms the other in six datasets. For example, although GAE performs better than GraphSage on Cora and CiteSeer datasets, MGAE-SAGE outperforms MGAE-GCN on these two datasets. It indicates that the best GNN encoder for MGAE varies on different graph scenarios. 
    \item Another promising property of MGAE we want to remark is that the results in Table~\ref{table_lp_planetoid} and~\ref{table_lp_ogb} are obtained under a high masking ratio ($\omega=0.7$), excepting ogbl-ppa. That is, we only feed 30\% original edges of the original graph to GNN encoder. Therefore, MGAE is naturally more efficient than traditional graph autoencoder models, since message propagation is the most time-consuming process of GNN models. Besides, it also indicates that a lot of node connections in graph data are redundant. This observation is consistent with the motivations for structure learning~\cite{franceschi2019learning,jin2020graph} or graph sparsification~\cite{zheng2020robust,wan2021edge}.  
    
\end{itemize}

\begin{table*}[t]
\centering
  \caption{Node classification performance on all datasets based on $70\%$ randomly edge masking. }
  \begin{small}
\setlength{\tabcolsep}{3.5pt}
  {
    \begin{tabular}{l c c c c c}
    \toprule
     Method &Cora &CiteSeer &PubMed &ogbn-arxiv &ogbn-proteins \\
      &ACC. &ACC. &ACC. &ACC. &AUC \\
     \midrule
    GCN &$83.60\pm0.52$ &$63.37\pm1.21$ &$78.23\pm1.63$ &$66.01 \pm 0.37$ &$61.67\pm0.35$\\
    GraphSage &$74.30\pm1.84$ &$60.20\pm2.15$ &$81.96 \pm 0.74$ &$64.79\pm2.91$ &$55.39\pm0.79$\\
    ARGVA &$85.86\pm0.72$ &$73.10\pm0.86$ &$81.85\pm1.01$ &$50.06\pm1.21$ &$40.73\pm0.68$\\
    SelfTask-GNN &$84.69\pm0.09$ &$71.82\pm0.13$ &$83.92\pm0.18$ &$68.30\pm0.02$ &$60.93\pm0.44$\\
    DGI &$85.41\pm0.34$ &$74.51\pm0.51$ &$85.95\pm0.66$ &$67.08\pm0.43$ &$50.31\pm0.55$\\
    GIC &$\bf{87.70\pm0.01}$ &$\bf{76.39\pm0.02}$ &$85.99\pm0.13$ &$64.00\pm0.22$ &$48.55\pm0.47$\\
    \midrule
    MGAE &$86.15\pm0.25$ &$74.60\pm0.06$ &$\bf{86.91\pm0.28}$ &$\bf{72.02\pm0.05}$ &$\bf{63.33\pm0.12}$\\
  \bottomrule
\end{tabular}}
\end{small}
\label{table_nc}
\end{table*}

\subsection{Node Classification}
In addition to link prediction, to further answer \textbf{Q1}, we evaluate our model on node classification over Planetoid (Cora, CiteSeer, and PubMed) and OGB (ogbn-arxiv and ogbn-protiens) datasets. We randomly split 10\% of all edges into validation set and use the remained 90\% edges as training set. The validation set is used to tune hyperparameters. After the model is trained, we use the full set of edges as input to generate node representations for downstream evaluation. Specifically, we train a SVM classifier on the learned node representations of all models, and apply 5-fold cross-validation to estimate the performance. To avoid randomness, we repeat the process for 10 times and report the average result in terms of Accuracy (ACC.) for Core, CiteSeer, PubMed and ogbn-arxiv and AUC for ogbn-proteins following~\cite{hu2021ogb,hu2020open}. We adopt the same baselines as link prediction settings, and report the result in Table~\ref{table_nc}. The major observations are made as follows. 
\begin{itemize}
    \item Our model MGAE performs consistently better than three graph-autoencoder based baselines (GCN, GraphSage, and ARGVA) across five datasets. Given that MGAE can obtain at least comparable results with classical graph autoencoders in link prediction task, it indicates that the proposed MGAE is a powerful graph autoencoder alternative. 
    \item Compared with self-supervised learning based models (DGI, GIC, and SelfTask-GNN), our model loses to the best performance of them on two small datasets (Cora and CiteSeer). However, MGAE outperforms them on three large datasets (PubMed, obgn-arxiv, and ogbn-proteins) by a large margin.
    \item SelfTask-GNN is a closely related work to MGAE, since they all focus on randomly masking some edges as self-supervised training task. However, according to the results in Table~\ref{table_lp_planetoid},~\ref{table_lp_ogb}, and~\ref{table_nc}, SelfTask-GNN does not perform well compared to state-of-the-art baselines due to the limited decoder design and masking strategies. These results show that our MGAE model is the first work that can successfully adopt self-supervised learning to boost the performance of classical graph autoencoders. 
\end{itemize}

\subsection{Sensitivity Analysis}
\begin{figure*}
\centering
\subfigure[PubMed]{
\includegraphics[width=6.5cm]{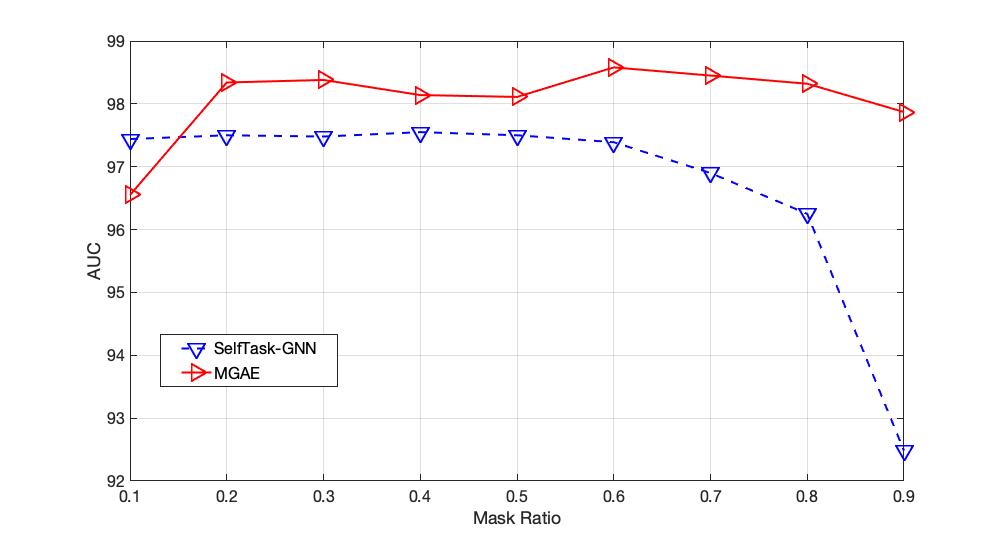}
}
\subfigure[ogbl-ddi]{
\includegraphics[width=6.5cm]{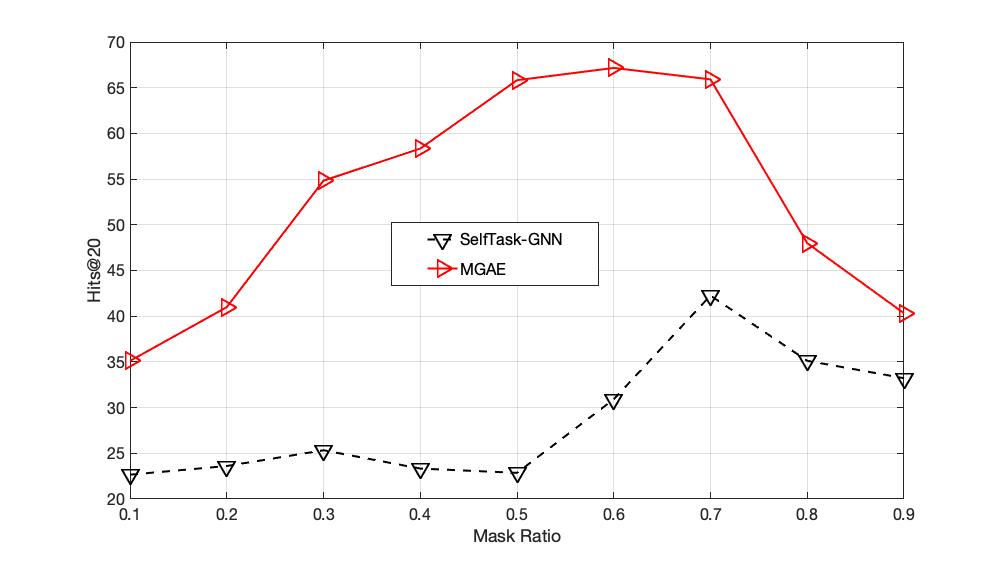}
}    
\caption{Performance of MGAE vs. SelfTask-GNN on link prediction with respect to $\omega$.}
\label{figure_sensi}
\vspace{-10pt}
\end{figure*}

In this section, we conduct experiments to verify the impacts of masking ratios $\omega$ on our model MGAE (\textbf{Q2}). To provide a comprehensive evaluation of MGAE, we include SelfTask-GNN for comparison since it can be viewed as a variant of our model by replacing the cross-correlation decoder with a simple MLP network. Specifically, we vary $\omega$ from 0.1 to 0.9 with a step size 0.1. Figure~\ref{figure_sensi} shows the results of MGAE-GCN and SelfTask-GNN on PubMed and ogbl-ddi datastes. Similar curves are observed on other datasets. From the figures, we have two major observations. 
\begin{itemize}
    \item The performance of MGAE first increases with the increasing of masking ratio $\omega$ until it reaches $0.7$, then it drops when $\omega$ further increases. Besides, our MGAE model performs relatively stable when $\omega$ is around 0.5 to 0.7. These results indicate the robustness of our model under high masking ratios. 
    \item MGAE performs consistently better than SelfTask-GNN on the two datasets across different $\omega$ values, except when $\omega<0.2$ on PubMed dataset. Besides, the performance gap is more significant when $\omega$ is around 0.5 and 0.7. These observations verify the effectiveness of our proposed cross-correlation decoder for self-supervised graph autoencoder training. 
\end{itemize}

\section{Conclusion}
We explore a novel masked graph autoencoder (MGAE) framework for unsupervised representation learning on graphs. It takes inspirations from self-supervised learning and can be viewed as a self-supervised graph autoencoder alternative. Different from vanilla graph autoencoder models, MGAE suggests to randomly mask a high proportion (i.e., 70\%) of the original graph structure as input and reconstruct only the masked edges for model training. Specifically, we introduce two edge masking strategies: undirected masking and directed masking, to generate valid self-supervisory tasks. Besides, we also propose a tailored cross-correlation decoder to effectively recover missing edges by capturing the cross representations between its head and tail nodes. Extensive experimental results across multiple open graph benchmarks validate the superiority of MGAE against state-of-the-art baselines in terms of link prediction and node classification tasks.

\appendix
\bibliographystyle{unsrt}
\bibliography{reference_nips}
\end{document}